\documentclass[sigconf, nonacm]{acmart}
\AtBeginDocument{%
  }

\setcopyright{acmlicensed}
\copyrightyear{2018}
\acmYear{2018}
\acmDOI{XXXXXXX.XXXXXXX}
\acmConference[Conference acronym 'XX]{Make sure to enter the correct
  conference title from your rights confirmation email}{June 03--05,
  2018}{Woodstock, NY}
\acmISBN{978-1-4503-XXXX-X/2018/06}

\usepackage{float}       
\usepackage{rotating}    
\usepackage{multirow}    
\usepackage{makecell}    
\usepackage[table]{xcolor}
\usepackage{url}
\usepackage{subfigure}
\usepackage[ruled,linesnumbered]{algorithm2e}
\usepackage{microtype}  
\usepackage{threeparttable}

\begin{document}

\title{COBRA++: Enhanced COBRA Optimizer with Augmented Surrogate Pool and Reinforced Surrogate Selection}


\author{Zipei Yu}
\email{zipei540@gmail.com}
\authornote{These authors contributed equally to this work.}
\orcid{0009-0007-2639-7981}
\affiliation{%
  \institution{South China University of Technology}
  \city{Guangzhou}
  \state{Guangdong}
  \country{China}
}

\author{Zhiyang Huang}
\email{scut.hzy@gmail.com}
\authornotemark[1]
\orcid{0009-0002-6499-0237}
\affiliation{%
  \institution{South China University of Technology}
  \city{Guangzhou}
  \state{Guangdong}
  \country{China}
}

\author{Hongshu Guo}
\email{guohongshu369@gmail.com}
\orcid{0000-0001-8063-8984}
\affiliation{%
  \institution{South China University of Technology}
  \city{Guangzhou}
  \state{Guangdong}
  \country{China}
}

\author{Yue-Jiao Gong}
\email{gongyuejiao@gmail.com}
\orcid{0000-0002-5648-1160}
\affiliation{%
  \institution{South China University of Technology}
  \city{Guangzhou}
  \state{Guangdong}
  \country{China}
}

\author{Zeyuan Ma}
\email{scut.crazynicolas@gmail.com}
\authornote{Corresponding author.}
\orcid{0000-0001-6216-9379}
\affiliation{%
  \institution{South China University of Technology}
  \city{Guangzhou}
  \state{Guangdong}
  \country{China}
}

\renewcommand{\shortauthors}{Yu et al.}

\begin{abstract}
The optimization problems in realistic world present significant challenges onto optimization algorithms, such as the expensive evaluation issue and complex constraint conditions. COBRA optimizer (including its up-to-date variants) is a representative and effective tool for addressing such optimization problems, which introduces 1) RBF surrogate to reduce online evaluation and 2) bi-stage optimization process to alternate search for feasible solution and optimal solution. Though promising, its design space, i.e., surrogate model pool and selection standard, is still manually decided by human expert, resulting in labor-intensive fine-tuning for novel tasks. In this paper, we propose a learning-based adaptive strategy (COBRA++) that enhances COBRA in two aspects: 1) An augmented surrogate pool to break the tie with RBF-like surrogate and hence enhances model diversity and approximation capability; 2) A reinforcement learning-based online model selection policy that empowers efficient and accurate optimization process. The model selection policy is trained to maximize overall performance of COBRA++ across a distribution of constrained optimization problems with diverse properties. We have conducted multi-dimensional validation experiments and demonstrate that COBRA++ achieves substantial performance improvement against vanilla COBRA and its adaptive variant. Ablation studies are provided to support correctness of each design component in COBRA++.     
\end{abstract}




\maketitle

\section{Introduction}
Constrained optimization problems arise widely in scientific and engineering disciplines,ranging from aerospace design and chemical process optimization to financial modeling and machine learning hyperparameter tuning \cite{Nocedal:2006jvv,COELLOCOELLO20021245,inbook}. The challenge in these problems lies in the complex feasible regions defined by constraints that are often non-convex or disconnected  \cite{zoutendijk1960methods,bazaraa2006nonlinear,boyd2004convex,polynomial}, substantially increasing the difficulty of locating global optima \cite{COELLOCOELLO20021245,article23,Coello2000}.
While many traditional optimization methods rely on repeated direct evaluation of the true objective and constraint functions, such approaches become impractical for real-world problems with expensive evaluations, where each function call may require costly simulations or intensive computations and the evaluation budget is severely limited \cite{conn2000trust,Regis2005ConstrainedGO,Liang2024ASO}. Consequently, there is a pressing demand for optimization techniques that can achieve high performance with a minimal number of true function evaluations, maximizing the utility of every acquired data point.

\begin{sloppypar}

Although numerous methods have been proposed for constrained optimization \cite{996017}, only a limited number of approaches are specifically designed for expensive constrained optimization \cite{book}. Among them, the \textbf{C}onstrained \textbf{O}ptimization \textbf{B}y \textbf{RA}dial basis function interpolation  (COBRA) \cite{RegisCOBRA} algorithm is widely recognized as a representative surrogate-assisted framework for expensive black-box constrained optimization \cite{Regis2005ConstrainedGO}. COBRA iteratively constructs computationally efficient surrogate models \cite{Vapnik1998StatisticalLT,ml} of the objective and constraint functions using Radial Basis Function (RBF) interpolation \cite{2003Radial} and guides the search with a bi-stage optimization process under a strict evaluation budget.
 


Despite its promising performance, COBRA can still be improved in two aspects: (1) COBRA typically employs a fixed surrogate model, which may not be well-suited for every problem landscapes, potentially leading to inaccurate approximations and reduced optimization performance. (2) COBRA's extension A-SACOBRA \cite{ASACOBRA} has shown that dynamically selecting different surrogate models for different problems can maintain low approximation errors, thereby improving the final optimization performance.
However, the strategy of selecting surrogate models in A-SACOBRA is solely based on a static, human-crafted heuristic, which still lacks flexibility and may not generalize well across diverse complex problem landscapes. Recently, Reinforcement Learning~(RL) is often used as an effective paradigm to learn a flexible strategy for parameter control and algorithm configuration \cite{metabbosurvey, ma2023metabox, ma2025metabox,DURGUT2021773,SHAO2025101817,Li2016LearningTO}. Inspired by such paradigm, in this paper, we propose COBRA++, an advanced framework for expensive constrained optimization problems. COBRA++ extends the vanilla COBRA by (1) significantly enriching the surrogate model pool with diverse approximation models to capture complex objective functions and intricate constraint boundaries, and (2) introducing an RL-based online surrogate selection policy to choose the most appropriate surrogate model at each step by leveraging current optimization features and model performance to enable cooperative surrogate utilization, thereby improving both optimization efficiency and solution quality. Extensive multi-dimensional experimental results demonstrate that our COBRA++ substantially outperforms both the vanilla COBRA and its variants. Additional ablation studies further verify the effectiveness of each proposed design component.
\end{sloppypar}
The remainder of this paper is organized as follows: Section \ref{sec:related works} presents the related works on constrained optimization, COBRA and RL-assisted optimization. Section \ref{sec:methodology} introduces the details of COBRA++. Section \ref{sec:experiment} is the experimental results with detailed analysis. Finally, we conclude our work in this paper in Section \ref{sec:conclusion}.
\section{Related Works}
\label{sec:related works}
\subsection{Constrained Optimization}
Constrained optimization problems are characterized by objective and constraint functions. Following the definition by Regis (2016) \cite{Regis2005ConstrainedGO}, a constrained black-box optimization problem can be formulated as:
\begin{equation}
\begin{aligned}
\min \quad & f(x) \\
\text{s.t.} \quad & g_i(x) \leq 0,\, i = 1,2,\dots,M \\
& \mathbf{a} \leq x \leq \mathbf{b}
\end{aligned}
\label{eq:constrained problem}
\end{equation}
where $x \in \mathbb{R}^d$ is the $d$-dimensional decision vector, $f(x)$ is the objective function, and $g_i(x)$ represents $M$ inequality constraints. Both $f$ and $g_i$ are deterministic, black-box functions.

Constrained optimization has been extensively studied, with representative approaches including penalty-based methods \cite{Fiacco1968NonlinearPS}, augmented Lagrangian methods \cite{Hestenes1969MultiplierAG,BERTSEKAS1982179}, feasible direction methods \cite{zoutendijk1960methods,bazaraa2006nonlinear}, and interior-point methods \cite{polynomial,boyd2004convex}. Evolutionary Algorithms (EAs), such as Genetic Algorithms \cite{10.7551/mitpress/1090.003.0004} and Particle Swarm Optimization \cite{488968}, have also been widely applied to constrained optimization problems with various constraint-handling techniques \cite{Coello2000,article23,996017}. 

In this paper, we consider a more challenging constrained optimization problem: expensive constrained optimization, where the objective function $f(x)$ and constraint functions $g_i(x)$ in Eq.~\ref{eq:constrained problem} are computationally expensive to evaluate. These problems are particularly challenging because the total evaluation budget is severely limited, making it difficult to obtain high-quality solutions and even to satisfy feasibility in some cases. To tackle such problems, one of the most representative approaches is COBRA, which first introduces surrogate models to ensure cheap function evaluations and employs a bi-stage optimization process to handle constraints and the objective function respectively. We elaborate the concepts and related works of COBRA in the next section.

\subsection{COBRA}

\begin{algorithm}[t]
\caption{COBRA-like Algorithm Framework}
\label{alg:cobra}
\KwIn {Initial sample size $N_0$, objective function $f(\cdot)$, constraint functions $g_i(\cdot)$,\, $i=1,\dots,M$}
\KwOut {Best feasible solution $x^*$}
\state {Initialize dataset $\mathcal{D} = \{(x_j, f(x_j), g_i(x_j))\}_{j=1}^{N_0}$}\\
\state {Train surrogate models $\hat f(x)$ and $\hat g_i(x), \, i=1,\dots,M$}\\
\While{\textnormal{evaluation budget not exhausted}}{
    \textbf{Stage 1: Constraint Satisfaction}\\
    \quad Solve surrogate subproblem to reduce constraint violations\\
    \textbf{Stage 2: Objective Improvement}\\
    \quad Solve surrogate subproblem to minimize $\hat f(x)$ under $\hat g_i(x) \le 0$\\
    Select a new point $x^*$ \\
    Evaluate true $f(x^*)$ and $g_i(x^*)$\\
    Update dataset $\mathcal{D} \leftarrow \mathcal{D} \cup \{(x^*, f(x^*), g_i(x^*))\}$\\
    Retrain surrogate models $\hat f(x)$ and $\hat g_i(x)$\\}
\KwRet Best feasible solution $x^*$
\end{algorithm}

\textbf{C}onstrained \textbf{O}ptimization \textbf{B}y \textbf{RA}dial basis function interpolation (COBRA) \cite{RegisCOBRA} is a surrogate-assisted framework specifically designed for expensive constrained black-box optimization problems. We summarize it in Algorithm~\ref{alg:cobra}. Here, $\hat f(x)$ and $\hat g_i(x)$ denote surrogate models that approximate the expensive objective and constraint functions $f(x)$ and $g_i(x)$, respectively. The vanilla COBRA constructs radial basis function (RBF) surrogate models for both the objective and constraint functions, enabling computationally cheap approximations of expensive evaluations. It iteratively refines these surrogates and employs a trust-region based search strategy to balance exploration and exploitation. At each iteration, COBRA alternates between constraint satisfaction and objective improvement, gradually steering candidate solutions toward the feasible region while optimizing the objective under strict evaluation budgets. This bi-stage optimization mechanism allows COBRA to effectively handle infeasible initial points and limited evaluation budgets, making it one of the most representative methods for expensive constrained optimization.

Addressing the lack of feasible solutions in some cases and the sensitivity of COBRA to extreme value distributions in complex landscapes, SACOBRA \cite{sacobra1,sacobra2,sacobra3} emerges as a key variant that enhances robustness and efficiency by introducing targeted improvements: it integrates a feasible solution repair strategy to recover near-feasible candidates that vanilla COBRA discards, and adopts the PLOG transformation to stabilize objective function scaling. Further extending this line of advancement, A-SACOBRA \cite{ASACOBRA} adds dynamic model selection capabilities to SACOBRA: instead of relying on a fixed RBF model, it adaptively switches between different RBF kernel types and parameter configurations based on real-time prediction error, enabling the surrogate to better align with the evolving optimization landscape. These successive refinements gradually mitigate the limitations of static strategies, though they still rely on heuristic-driven adjustments rather than learning-based decision-making. Nevertheless, the effectiveness of COBRA-like algorithms largely depends on manually designed surrogate management strategies, which motivates us to explore learning-based approaches for more adaptive optimization.

\subsection{RL-Assisted Optimization}
Reinforcement Learning~(RL), with its ability to learn optimal policies through interaction with an environment, has proven to be a powerful tool for automating various aspects of optimization \cite{ma2023metabox,ma2025metabox, metabbosurvey}. Recent studies have applied RL to a wide range of optimization problems, including single-objective optimization \cite{guo2025designx,guo2025advancing,guo2025rldeafl,ma2025surrogate,gleet,rl-das,symbol,offlineqlearning,yimeilibog,yimeisoo}, multi-objective optimization \cite{LImoo,feimoo,tianyemoo,hanmoo,xuemoo,huangmoo}, multimodal optimization \cite{rlemmo,adpmmo,xiammo,HONGmmo,sunmmo}, and multitask optimization \cite{metamto,wenyingongmto,mtosurvey,Martinezmto,kctanmto}. Moreover, RL has also been explored for more challenging scenarios such as constrained optimization \cite{DURGUT2021773,wenyingongcop,wenyingongcmop1,wenyingongcmop2}. However, despite these advances, the application of RL to expensive constrained optimization remains underexplored, where the strict evaluation budget and complex feasibility requirements pose significant challenges for learning effective control policies.

Therefore, in this work, we build upon COBRA as a foundational framework and enhance it with a elaborately designed surrogate model pool and an RL-assisted mechanism for dynamic surrogate selection. This framework improves the flexibility of COBRA in tackling expensive constrained optimization problems, enabling adaptive surrogate utilization throughout the optimization process and ultimately leading to improved overall optimization performance.

\section{Methodology}
\label{sec:methodology}
\subsection{Overview}
\label{subsec:overview}
\begin{figure}[t]
    \centering
    \includegraphics[width=0.9\columnwidth]{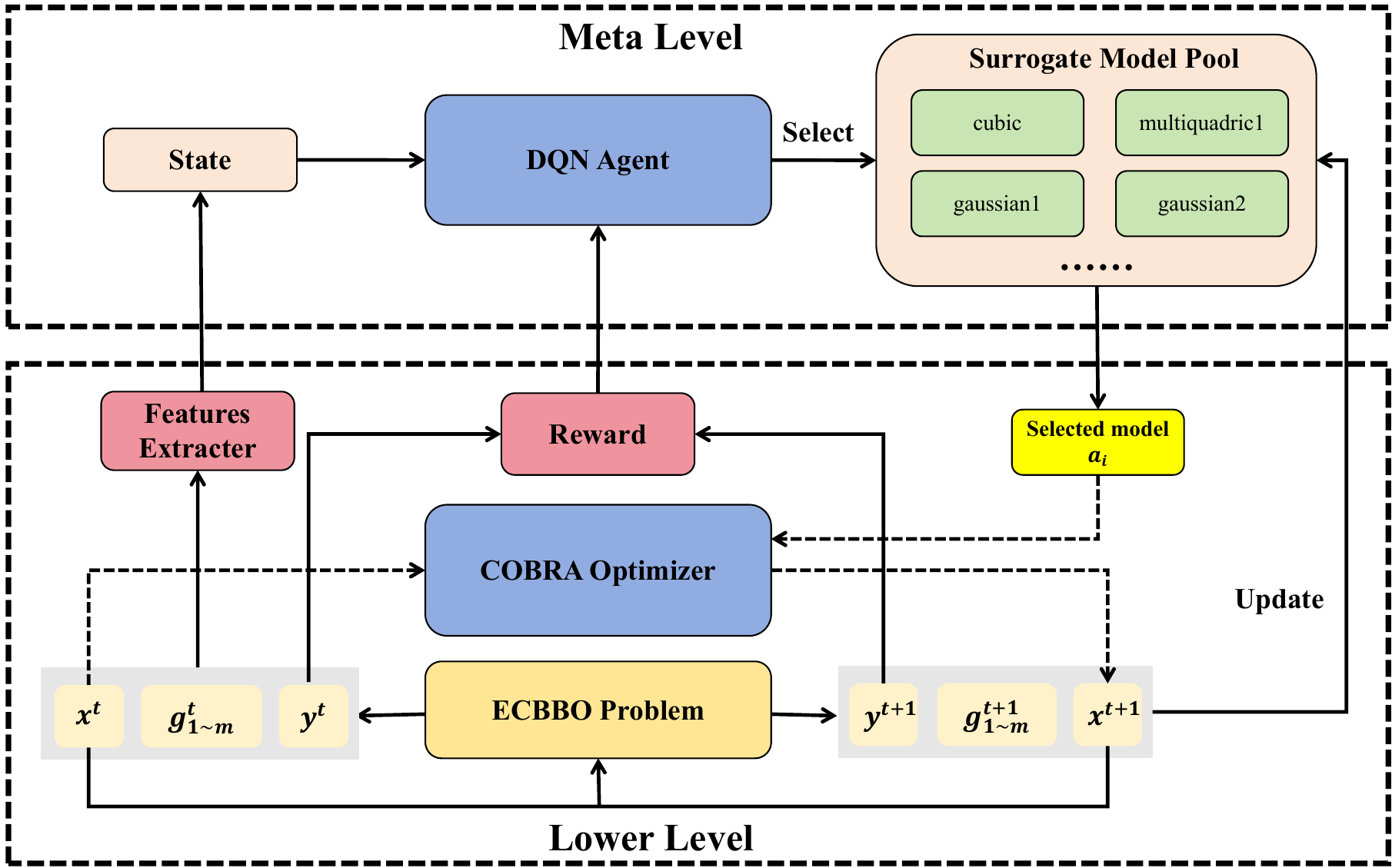}
    \caption{Overview of COBRA++.}
    \label{fig:overview} 
\end{figure}

\begin{sloppypar}
Our COBRA++ enhances the vanilla COBRA with two major contributions. Firstly, we expand the surrogate model pool, thereby increasing model diversity and overall approximation capability. Secondly, we propose an RL-driven surrogate selection policy. The learned policy dynamically determines which surrogate model should guide the optimization at each time step, enabling more adaptive and effective optimization control. 
As illustrated in Fig.~\ref{fig:overview}, COBRA++ follows a bi-level architecture. At the lower level, the COBRA optimizer utilizes the surrogate model selected by the RL agent to improve the solution from $x^t$ to $x^{t+1}$, which is then evaluated by the expensive objective and constraint functions. The evaluation results are used to compute the reward, update the optimization state, and retrain all surrogate models in the pool. At the meta level, the RL agent receives the current optimization state and reward, then dynamically selects a surrogate model at each time step, which is modeled as a Markov Decision Process~(MDP). The agent is trained to maximize the accumulated reward across a distribution of expensive constrained optimization problems. The details of surrogate model pool, MDP design and the overall workflow will be provided in Section~\ref{subsec:pool}, Section~\ref{subsec:mdp}, Section~\ref{subsec:workflow} respectively.
\end{sloppypar}



\subsection{Surrogate Model Pool}
\label{subsec:pool}

The surrogate model pool adopts \textbf{Radial Basis Function (RBF)} networks \cite{2003Radial} as the base model, leveraging their strong ability to approximate nonlinear functions and support local interpolation. Each RBF model is defined as:
\begin{equation}
s(x) = \sum_{k=1}^K \lambda_k \cdot \phi\left(\frac{\|x - c_k\|}{\sigma}\right) + \sum_{t=0}^T \beta_t \cdot p_t(x)
\end{equation}
where \textbf{$\phi(\cdot)$} is the RBF kernel function, 
\textbf{$c_k$} is the center of the $k$-th RBF kernel, \textbf{$\sigma$} is the kernel width based on sample distribution, \textbf{$\lambda_k$} is the weight of the $k$-th RBF kernel, \textbf{$p_t(x)$}is the polynomial basis function of degree $t$, and maximum degree $T$ is determined by sample size, \textbf{$\beta_t$} is the weight of the $t$-th polynomial basis.

COBRA++ maintains a pool of 11 radial basis function (RBF) surrogate models with different kernel types and width factors $w$.
The pool consists of one cubic RBF, five multiquadric RBFs with width factors $w$:$[0.01, 0.2, 0.5, 1, 5]$, and five Gaussian RBFs with width factors $w$:$[0.01, 0.2, 0.5, 1, 5]$. These kernels are defined in Table~\ref{tab:RBF}. All surrogate models in the pool are treated equally and are selected dynamically during optimization.

\begin{table}[t]
\caption{Different RBF kernels used in COBRA++}
\label{tab:RBF}
\resizebox{0.4\columnwidth}{!}{%
\begin{tabular}{c|c}
\hline
Function name&$\phi(r)$\\ \hline
Cubic&$r^3$   \\ 
Multiquadric& $\sqrt{1+\left( \frac{r}{w} \right)^2}$\\
Gaussian&$e^{-\left( \frac{r}{w} \right)^2}$ \\ \hline
\end{tabular}
}
\end{table}






\subsection{MDP Design}
\label{subsec:mdp}
We formulate the surrogate model selection process as a Markov Decision Process (MDP). In specific, at each optimization time step $t$, the agent observes the current surrogate model and optimization state $s^t \in S$, and selects a surrogate model $a^t \in A$. Given the selected surrogate model, the COBRA optimizer updates the solution from $x^t$ to $x^{t+1}$ and evaluates it using the expensive objective and constraint functions, resulting in a new state $s^{t+1}$.
The agent receives a reward $r^t$, which reflects the optimization progress. We next describe the state design, action space design, reward design, and network architecture in the following subsections.



\begin{table*}[h]
\renewcommand{\arraystretch}{1.5}
  \caption{State Formulations and Detailed Descriptions}
  \label{tab:state_features}
  \begin{tabular}{cccp{7cm}}
  \toprule
  Class & Feature ID & Formula & \hspace{3cm}Description \\
  \hline
  \multirow{4}{*}[-60pt]{\shortstack{Surrogate\\Feature\\$s_{m,i}^t$}}  
  &\multirow{1}{*}[-10pt]{$1$} & \multirow{1}{*}[-8pt]{$\frac{1}{|\mathcal{D}|} \sum_{j=1}^{|\mathcal{D}|} \frac{|\hat f(x_j) - f(x_j)|}{\max_{\mathcal{D}}f(x) - \min_{\mathcal{D}}f(x)}$} 
  & Average prediction error of the surrogate model, normalized by the range of the objective values in current dataset $\mathcal{D}$\\ 
  \cline{2-4} 
  & \multirow{1}{*}[-16pt]{$2 \sim 6$} &  \multirow{1}{*}[-10pt]{$\begin{cases}1, & \text{if model } a_i \text{ selected at step } t-k \\ 0, & \text{otherwise}\end{cases} (k=1, ... ,5)$ }
  & Binary vector indicating the usage history of the surrogate model over the last 5 optimization steps, where 1 indicates model selection at a given step, and 0 otherwise. \\ 
  \cline{2-4} 
   & \multirow{1}{*}[-10pt]{$7$} & \multirow{1}{*}[-8pt]{$\frac{N_{succ}}{5}$}
   & An optimization step is successful if the constraints are fulfilled by using $a_i$ in COBRA++. $N_{succ}$ is the number of successful steps in the last five steps where $a_i$ is used. \\ 
   \cline{2-4} 
  & \multirow{1}{*}[-16pt]{$8$} & \multirow{1}{*}[-12pt]{$\frac{\sum_{k=0}^t \mathbb{I}(f(x^{k}) < f(x^{k-1})) \cdot \mathbb{I}(a_i \text{ selected at step } k)}{\sum_{i'=1}^{11}\sum_{k=0}^t \mathbb{I}(f(x^{k}) < f(x^{k-1})) \cdot \mathbb{I}(a_{i'} \text{ selected at step } k)}$ }
  & Normalized cumulative contribution of the surrogate model $a_i$, defined as the proportion of objective improvements attributed to $a_i$ among all surrogate models up to step $t$.\\ 
  \midrule
  \multirow{2}{*}[-5pt]{\shortstack{Global\\Feature\\$s_g^t$}}
  & \multirow{1}{*}[-6pt]{$1$} & \multirow{1}{*}[-6pt]{$ \text{std}(f(x^{t-4}), f(x^{t-3}), f(x^{t-2}), f(x^{t-1}), f(x^t))$} 
  & Standard deviation of the objective values from the last 5 optimization steps \\ 
  \cline{2-4} 
  & $2$ & $\frac{FEs}{MaxFEs}$ 
  & Ratio of consumed function evaluations \\
  \bottomrule
\end{tabular}
\end{table*}
\subsubsection{State Design}
\begin{sloppypar}
\label{sec:state}
Previous RL-assisted optimization studies mainly focus on describing the optimization process itself, such as solution information or objective improvement. 
Since COBRA++ is a model selection framework, we design a novel state representation that captures not only global optimization features but also, more importantly, surrogate specific features. Specifically, at time step $t$, the state is defined as
$s^t = \{\{ s_{m,i}^t \}_{i=1}^{11}, s_g^t\}$, where $s_{m,i}^t \in \mathbb{R}^8$ denotes the feature vector associated with the $i$-th surrogate model, and $s_g^t \in \mathbb{R}^2$ represents the global optimization features. The detailed definitions and descriptions of these state components are summarized in Table~\ref{tab:state_features}, where $\mathbb{I}(condition)$ is an indicator function that equals 1 if the condition is true and 0 otherwise. This state design profiles historical performance metrics of each surrogate model in the model pool and the current optimization progress, ensuring more informed and corresponding adaptive surrogate selection decisions.
\end{sloppypar}
\subsubsection{Action Space Design}
As there are 11 surrogate models in the model pool, the action space $A$ comprises 11 optional actions, where each action $a^t_i \in A$ corresponds to selecting the $i$-th surrogate model to guide the COBRA optimizer at optimization step $t$. In this way, the RL agent directly controls the surrogate selection strategy, enabling adaptive model utilization throughout the optimization process.




\subsubsection{Reward Design}
\label{sec:reward}
We utilize a simple reward function in COBRA++:
    \begin{equation}
    r^t = \begin{cases} 
    1, & \text{if } f(x^{t+1}) < f(x^t) \text{ and } \forall m, g_m(x^{t+1}) \leq 0 \\
    0, & \text{otherwise}
    \end{cases}
    \end{equation}
 where $x^t$ and $x^{t+1}$ are the solutions at time step $t$ and $t+1$ respectively. If a better and feasible solution is found, a positive reward signal will be returned to the RL agent. This reward directly links model selection to optimization progress, ensuring the agent learns an effective policy.

\subsubsection{Network Architecture}
\label{sec:architecture}
\begin{figure}[t]
    \centering
    \includegraphics[width=0.9\columnwidth]{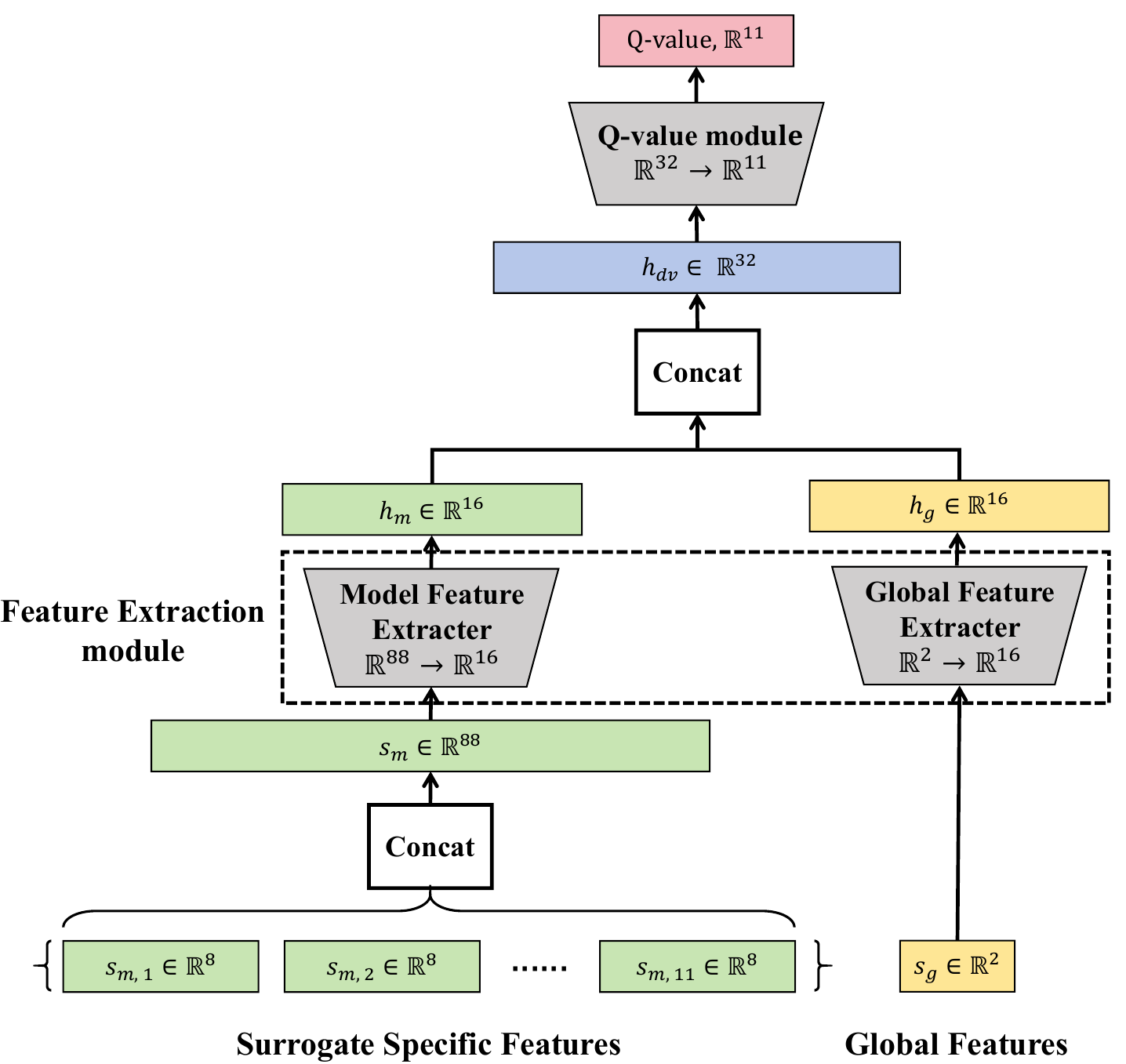}
    \caption{Network Architecture of COBRA++.}
    \label{fig:architecture} 

\end{figure}

COBRA++ adopts a Deep Q-Network architecture, composed of two important modules: a feature extraction module and a Q-value module. We depict the network architecture of COBRA++ in Figure~\ref{fig:architecture}.
The feature extraction module contains two separate MLPs to process different state components. Specifically, the surrogate-specific features $\{s_{m,i}\}_{i=1}^{11}$ are concatenated and fed into a model feature extractor, which maps the high-dimensional features into a hidden model state vector $h_m \in \mathbb{R}^{16}$. The global optimization features $s_g$ are processed by a global feature extractor, producing a hidden global state vector $h_g \in \mathbb{R}^{16}$. Then, $h_m$ and $h_g$ are concatenated to form decision vector $h_{dv} = [h_m, h_g] \in \mathbb{R}^{32}$.
The Q-value module is an MLP that takes $h_{dv}$ as input and outputs an 11-dimensional Q-value vector, where each entry corresponds to the estimated Q-value of selecting a specific surrogate model.





\subsection{Overall Workflow of COBRA++}
\label{subsec:workflow}

\begin{algorithm}[t]
\caption{COBRA++ Optimization Framework.}
\label{alg:cobra++}\small
\KwIn{Training Problem Set $P_{train}$, Initialized selection policy $\pi_\theta$, Initialized surrogate pool $\mathcal{P}$, Initial sample size $N_0$, Maximum training epoch $MaxEpoch$, Maximum function evaluations $MaxFEs$ per optimization run, COBRA optimizer $\mathcal{O}$.}
\KwOut{Optimal selection policy $\pi_{\theta^*}$.}
\state{Initialize replay buffer $RB \leftarrow \emptyset$;\\}
\For{$epoch \leftarrow 0$ \KwTo $MaxEpoch$}{
    \For{\textnormal{each problem $f$ in } $P_{train}$}{
        \state {Initialize dataset $\mathcal{D} = \{(x_j, f(x_j), g_i(x_j))\}_{j=1}^{N_0}$\\}
        \state{Train models in $\mathcal{P}$ with $\mathcal{D}$\\}
        \state{Set time flag $t \leftarrow 0$, function evaluations $FEs \leftarrow N_0$\\}
        \While{$FEs \leq MaxFEs$}{
            \state{Extract states $s^{t}$ following Section~\ref{sec:state}\\}
            \state{Sample a surrogate model $a^t \sim \pi_\theta(s^t)$ \\}
            \state{$\mathcal{O}$ executes line 4--11 in Algorithm~\ref{alg:cobra} with $a^t$\\}
            \state{Compute reward $r^t$ following Section~\ref{sec:reward}\\}
            \state{Extract state $s^{t+1}$ following Section~\ref{sec:state}\\}
            \state{Insert $(s^t, a^t, r^t, s^{t+1})$ in $RB$\\}
            \state{Update $\pi_\theta$ following Equation~ \ref{eq:loss}\\}
            \state{$t \leftarrow t+1$, $FEs \leftarrow FEs + 1$\\}
        }
    }
}
\KwRet{\textnormal{The trained model selection policy} $\pi_{\theta^*}$}
\end{algorithm}

\begin{sloppypar}
We present the pseudo code of the overall workflow of COBRA++ in Algorithm~\ref{alg:cobra++}. We first initialize a policy $\pi_\theta$, whose network architecture is introduced in Section~\ref{sec:architecture}. 
During the lower-level optimization process (line~7--15), for each training problem in $P_{train}$, an initial set of candidate solutions is first generated and evaluated using the expensive objective and constraint functions. These evaluated samples are then used to train all 11 surrogate models in the model pool $\mathcal{P}$. At each optimization step $t$, the current state $s^t$ is constructed, and a surrogate model $a^t$ is selected according to the policy $\pi_\theta$. Then, the COBRA optimizer $\mathcal{O}$ performs one optimization step (line~4--11 in Algorithm~\ref{alg:cobra}) using the selected surrogate model to generate new candidate solutions. Subsequently, the next state $s^{t+1}$ and the reward $r^t$ are computed and the transition $(s^t, a^t, r^t, s^{t+1})$ is stored in the replay buffer $RB$. To train the Q-network, transitions are sampled from $RB$ and the network parameters are updated by minimizing the Bellman loss: 
\end{sloppypar}
\begin{equation}\label{eq:loss}
   \mathcal{L}(\theta) = \mathbb{E}_{(s^t,a^t,r^t,s^{t+1}) \sim RB} \left[ \left( y^t - \pi_\theta(s^t, a^t) \right)^2 \right], 
\end{equation}
where 
\begin{equation}
    y^t=r^t + \gamma \mathop{\arg\max}\limits_{a \in A} \pi_\theta(s^{t+1}, a),
\end{equation}

$\gamma$ is the discount factor. The network parameters $\theta$ are updated periodically with a fixed training interval.

Here we provide two important notes: (1) Unlike vanilla COBRA, which updates only the surrogate model used in the current iteration, COBRA++ retrains all 11 surrogate models in the pool using the newly evaluated and the previously used data points, ensuring consistent and up-to-date model predictions. (2) An $\epsilon$-greedy strategy is employed for model selection during training, while the model is selected greedily during testing.


\begin{table*}[h]
\centering
\caption{Performance Comparison on 10D Problems}
\label{tab:10d_performance}
\footnotesize
\resizebox{\textwidth}{!}{%
\begin{tabular}{c|*{4}{c|}|*{4}{c|}|*{3}{c|}c}
\hline
\multirow{3}{*}{}
& \multicolumn{4}{c||}{100 evaluations}
& \multicolumn{4}{c||}{150 evaluations}
& \multicolumn{4}{c}{200 evaluations} \\
\cline{2-5} \cline{6-9} \cline{10-13}
& COBRA & SACOBRA & A-SACOBRA & COBRA++
& COBRA & SACOBRA & A-SACOBRA & COBRA++
& COBRA & SACOBRA & A-SACOBRA & COBRA++ \\
\cline{2-13}
& mean  & mean  & mean  & mean 
& mean  & mean  & mean  & mean 
& mean  & mean  & mean  & mean 
\\ & $\pm$(std) &  $\pm$(std) & $\pm$(std) & $\pm$(std) &  $\pm$(std) &  $\pm$(std) & $\pm$(std) & $\pm$(std) & $\pm$(std) & $\pm$(std) & $\pm$(std) & $\pm$(std) 
\\ \hline
\multirow{2}{*}{Sphere1} & 1.68E+01 & 2.90E+01 & \cellcolor{gray!30}\textbf{4.13E+01} & 4.04E+01 & 2.92E+01 & 4.17E+01 & 6.41E+01 & \cellcolor{gray!30}\textbf{4.15E+02} & 2.92E+01 & 4.17E+01 & 6.41E+01 & \cellcolor{gray!30}\textbf{1.89E+02} \\
 & $\pm$(4.24E+00) & $\pm$(9.98E+00) & $\pm$(2.34E+00) & $\pm$(1.56E+01) & $\pm$(1.01E+01) & $\pm$(2.39E+00) & $\pm$(2.00E+01) & $\pm$(2.48E+02) & $\pm$(1.01E+01) & $\pm$(2.38E+00) & $\pm$(2.00E+01) & $\pm$(3.89E+01) \\ \hline
\multirow{2}{*}{Sphere2} & 2.50E+00 & 2.84E+00 & 2.98E+00 & \cellcolor{gray!30}\textbf{8.61E+00} & 2.88E+00 & 2.82E+00 & 2.96E+00 & \cellcolor{gray!30}\textbf{1.82E+01} & 2.84E+00 & 2.96E+00 & 3.04E+00 & \cellcolor{gray!30}\textbf{2.94E+01} \\
 & $\pm$(2.30E-01) & $\pm$(1.20E-01) & $\pm$(2.00E-02) & $\pm$(5.07E+00) & $\pm$(2.10E-01) & $\pm$(1.50E-01) & $\pm$(1.00E-02) & $\pm$(1.42E+01) & $\pm$(1.00E-01) & $\pm$(1.00E-02) & $\pm$(5.00E-02) & $\pm$(2.56E+01) \\ \hline
\multirow{2}{*}{Sphere3} & 5.80E+00 & 6.71E+00 & \cellcolor{gray!30}\textbf{8.26E+00} & 7.39E+00 & 8.22E+00 & 1.06E+01 & 1.03E+01 & \cellcolor{gray!30}\textbf{1.42E+02} & 8.42E+00 & 1.07E+01 & 1.12E+01 & \cellcolor{gray!30}\textbf{1.43E+01} \\
 & $\pm$(6.30E-01) & $\pm$(2.90E-01) & $\pm$(1.25E+00) & $\pm$(2.30E+00) & $\pm$(1.24E+00) & $\pm$(3.65E+00) & $\pm$(3.93E+00) & $\pm$(1.15E+02) & $\pm$(1.44E+00) & $\pm$(2.96E+00) & $\pm$(4.08E+00) & $\pm$(2.39E+00) \\ \hline
\multirow{2}{*}{Sphere4} & 6.51E+01 & 8.08E+01 & 9.88E+01 & \cellcolor{gray!30}\textbf{1.03E+02} & 7.91E+01 & 1.35E+02 & \cellcolor{gray!30}\textbf{1.35E+02} & 1.35E+02 & 7.91E+01 & 1.35E+02 & 1.35E+02 & \cellcolor{gray!30}\textbf{1.46E+02} \\
 & $\pm$(4.27E+01) & $\pm$(2.69E+01) & $\pm$(8.98E+00) & $\pm$(4.90E+00) & $\pm$(5.55E+01) & $\pm$(1.00E-02) & $\pm$(1.00E-04) & $\pm$(1.00E-04) & $\pm$(5.55E+01) & $\pm$(1.00E-02) & $\pm$(1.00E-04) & $\pm$(2.00E-04) \\ \hline
\multirow{2}{*}{Sphere5} & 2.00E+02 & 1.49E+04 & 1.49E+04 & \cellcolor{gray!30}\textbf{2.38E+04} & 2.00E+02 & 1.49E+04 & 1.49E+04 & \cellcolor{gray!30}\textbf{1.51E+04} & 2.00E+02 & 1.49E+04 & 1.49E+04 & \cellcolor{gray!30}\textbf{1.57E+04} \\
 & $\pm$(7.02E+00) & $\pm$(1.47E+04) & $\pm$(1.48E+04) & $\pm$(6.61E+03) & $\pm$(7.01E+00) & $\pm$(1.47E+04) & $\pm$(1.48E+04) & $\pm$(1.50E+04) & $\pm$(7.01E+00) & $\pm$(1.47E+04) & $\pm$(1.48E+04) & $\pm$(1.55E+04) \\ \hline
\multirow{2}{*}{Sphere6} & 1.23E+03 & 1.05E+04 & \cellcolor{gray!30}\textbf{4.63E+05} & 1.18E+05 & 1.26E+03 & 1.05E+04 & \cellcolor{gray!30}\textbf{4.63E+05} & 1.07E+05 & 1.26E+03 & 1.05E+04 & 4.63E+05 & \cellcolor{gray!30}\textbf{2.16E+06} \\
 & $\pm$(1.21E+03) & $\pm$(8.08E+03) & $\pm$(4.44E+05) & $\pm$(1.06E+05) & $\pm$(1.17E+03) & $\pm$(8.08E+03) & $\pm$(4.44E+05) & $\pm$(2.60E+03) & $\pm$(1.17E+03) & $\pm$(8.08E+03) & $\pm$(4.44E+05) & $\pm$(2.15E+06) \\ \hline
\multirow{2}{*}{Ellipsoid1} & 2.55E+01 & 4.41E+01 & 4.23E+01 & \cellcolor{gray!30}\textbf{5.08E+01} & 4.49E+01 & 6.02E+01 & 1.77E+02 & \cellcolor{gray!30}\textbf{2.28E+02} & 5.64E+01 & 1.87E+02 & \cellcolor{gray!30}\textbf{2.97E+02} & 2.00E+02 \\
 & $\pm$(1.77E+00) & $\pm$(1.68E+01) & $\pm$(1.86E+01) & $\pm$(1.93E+01) & $\pm$(1.46E+01) & $\pm$(7.10E-01) & $\pm$(1.66E+02) & $\pm$(1.36E+02) & $\pm$(2.50E+01) & $\pm$(1.05E+02) & $\pm$(4.54E+00) & $\pm$(2.34E+00) \\ \hline
\multirow{2}{*}{Ellipsoid2} & 2.62E+01 & 2.24E+01 & 1.75E+01 & \cellcolor{gray!30}\textbf{4.09E+01} & 2.62E+01 & 2.24E+01 & 1.75E+01 & \cellcolor{gray!30}\textbf{8.43E+01} & 2.62E+01 & 4.64E+01 & 4.15E+01 & \cellcolor{gray!30}\textbf{5.92E+01} \\
 & $\pm$(1.28E+00) & $\pm$(5.09E+00) & $\pm$(1.50E-01) & $\pm$(2.87E+01) & $\pm$(1.28E+00) & $\pm$(5.08E+00) & $\pm$(1.50E-01) & $\pm$(3.63E+01) & $\pm$(1.28E+00) & $\pm$(1.89E+01) & $\pm$(2.39E+01) & $\pm$(6.45E+00) \\ \hline
\multirow{2}{*}{Ellipsoid3} & 1.02E+01 & \cellcolor{gray!30}\textbf{1.54E+01} & 1.20E+01 & 8.11E+00 & 1.04E+01 & 1.75E+01 & 1.39E+01 & \cellcolor{gray!30}\textbf{3.18E+01} & 1.08E+01 & \cellcolor{gray!30}\textbf{1.75E+01} & 1.39E+01 & 1.56E+01 \\
 & $\pm$(8.71E+00) & $\pm$(3.49E+00) & $\pm$(1.30E-01) & $\pm$(4.18E+00) & $\pm$(8.94E+00) & $\pm$(1.83E+00) & $\pm$(1.75E+00) & $\pm$(1.78E+01) & $\pm$(8.50E+00) & $\pm$(1.83E+00) & $\pm$(1.75E+00) & $\pm$(1.07E+00) \\ \hline
\multirow{2}{*}{Ellipsoid4} & \cellcolor{gray!30}\textbf{1.51E+01} & 3.98E+00 & 8.39E+00 & 3.98E+00 & \cellcolor{gray!30}\textbf{1.87E+01} & 3.98E+00 & 8.83E+00 & 5.23E+00 & 1.87E+01 & 5.96E+00 & 8.86E+00 & \cellcolor{gray!30}\textbf{2.16E+01} \\
 & $\pm$(1.28E+01) & $\pm$(1.61E+00) & $\pm$(2.79E+00) & $\pm$(2.98E+00) & $\pm$(1.63E+01) & $\pm$(1.61E+00) & $\pm$(3.22E+00) & $\pm$(3.87E+00) & $\pm$(1.63E+01) & $\pm$(1.21E+00) & $\pm$(3.26E+00) & $\pm$(3.76E+00) \\ \hline
\multirow{2}{*}{Ellipsoid5} & 1.47E+00 & 1.18E+00 & 2.72E+00 & \cellcolor{gray!30}\textbf{1.38E+01} & 2.72E+00 & 1.88E+00 & \cellcolor{gray!30}\textbf{3.25E+00} & 3.14E+00 & 3.28E+00 & 2.45E+00 & 3.25E+00 & \cellcolor{gray!30}\textbf{3.66E+00} \\
 & $\pm$(4.70E-01) & $\pm$(1.80E-01) & $\pm$(1.37E+00) & $\pm$(1.16E+01) & $\pm$(1.36E+00) & $\pm$(5.30E-01) & $\pm$(8.30E-01) & $\pm$(7.50E-01) & $\pm$(8.00E-01) & $\pm$(3.00E-02) & $\pm$(8.30E-01) & $\pm$(9.20E-01) \\ \hline
\multirow{2}{*}{Ellipsoid6} & 1.02E+00 & 9.90E-01 & 1.12E+00 & \cellcolor{gray!30}\textbf{1.22E+00} & 1.04E+00 & 1.21E+00 & 1.29E+00 & \cellcolor{gray!30}\textbf{1.31E+00} & 1.27E+00 & 1.29E+00 & 1.42E+00 & \cellcolor{gray!30}\textbf{2.82E+00} \\
 & $\pm$(2.00E-02) & $\pm$(4.00E-02) & $\pm$(1.70E-01) & $\pm$(1.60E-01) & $\pm$(8.00E-02) & $\pm$(8.00E-02) & $\pm$(1.00E-02) & $\pm$(2.00E-02) & $\pm$(1.00E-02) & $\pm$(1.00E-02) & $\pm$(1.30E-01) & $\pm$(1.40E+00) \\ \hline
\multirow{2}{*}{Bent Cigar1} & 3.57E+00 & 3.80E+00 & 3.91E+00 & \cellcolor{gray!30}\textbf{8.06E+00} & 3.72E+00 & 4.99E+00 & \cellcolor{gray!30}\textbf{6.96E+00} & 4.29E+00 & 4.15E+00 & 5.39E+00 & 7.72E+00 & \cellcolor{gray!30}\textbf{1.08E+01} \\
 & $\pm$(3.00E-02) & $\pm$(2.10E-01) & $\pm$(9.00E-02) & $\pm$(2.83E+00) & $\pm$(1.20E-01) & $\pm$(1.13E+00) & $\pm$(8.20E-01) & $\pm$(1.10E-01) & $\pm$(2.60E-01) & $\pm$(7.10E-01) & $\pm$(1.58E+00) & $\pm$(6.10E-01) \\ \hline
\multirow{2}{*}{Bent Cigar2} & 2.96E+00 & 5.77E+00 & 9.85E+00 & \cellcolor{gray!30}\textbf{1.45E+01} & 4.99E+00 & 8.99E+00 & 1.30E+01 & \cellcolor{gray!30}\textbf{2.68E+01} & 1.39E+01 & 2.50E+01 & \cellcolor{gray!30}\textbf{2.94E+01} & 2.57E+01 \\
 & $\pm$(1.90E-01) & $\pm$(2.62E+00) & $\pm$(1.46E+00) & $\pm$(7.09E+00) & $\pm$(1.54E+00) & $\pm$(2.44E+00) & $\pm$(1.60E+00) & $\pm$(1.53E+00) & $\pm$(1.05E+01) & $\pm$(4.90E-01) & $\pm$(4.00E+00) & $\pm$(7.10E-01) \\ \hline
\multirow{2}{*}{Bent Cigar3} & 1.96E+00 & 7.63E+00 & \cellcolor{gray!30}\textbf{1.43E+01} & none & 4.01E+00 & 9.75E+00 & \cellcolor{gray!30}\textbf{1.76E+01} & 8.70E-01 & 5.35E+00 & 1.11E+01 & \cellcolor{gray!30}\textbf{2.28E+01} & 1.47E+01 \\
 & $\pm$(9.60E-01) & $\pm$(4.69E+00) & $\pm$(2.01E+00) & - & $\pm$(3.16E+00) & $\pm$(2.58E+00) & $\pm$(5.23E+00) & $\pm$(1.00E-01) & $\pm$(4.50E+00) & $\pm$(1.23E+00) & $\pm$(1.04E+01) & $\pm$(2.87E+00) \\ \hline
\multirow{2}{*}{Bent Cigar4} & none & 1.13E+00 & \cellcolor{gray!30}\textbf{3.66E+00} & none & none & 1.13E+00 & \cellcolor{gray!30}\textbf{7.76E+00} & none & 3.40E+00 & 3.47E+00 & \cellcolor{gray!30}\textbf{8.97E+00} & 2.00E+00 \\
 & - & $\pm$(1.30E-01) & $\pm$(2.39E+00) & - & - & $\pm$(1.30E-01) & $\pm$(6.49E+00) & - & $\pm$(2.62E-01) & $\pm$(3.90E-01) & $\pm$(5.32E+00) & $\pm$(9.95E-01) \\ \hline
\multirow{2}{*}{Bent Cigar5} & 2.13E+00 & 2.56E+00 & 2.69E+00 & \cellcolor{gray!30}\textbf{7.75E+00} & 2.13E+00 & 2.56E+00 & \cellcolor{gray!30}\textbf{2.69E+00} & none & 2.13E+00 & 2.56E+00 & 2.69E+00 & \cellcolor{gray!30}\textbf{7.75E+00} \\
 & $\pm$(1.80E-01) & $\pm$(2.60E-01) & $\pm$(1.30E-01) & $\pm$(3.57E+00) & $\pm$(1.70E-01) & $\pm$(2.60E-01) & $\pm$(1.30E-01) & - & $\pm$(1.70E-01) & $\pm$(2.60E-01) & $\pm$(1.30E-01) & $\pm$(3.57E+00) \\ \hline
\multirow{2}{*}{Bent Cigar6} & 1.47E+00 & \cellcolor{gray!30}\textbf{7.62E+00} & 1.81E+00 & 6.96E+00 & 7.62E+00 & \cellcolor{gray!30}\textbf{7.93E+00} & 2.14E+00 & 3.56E+00 & 7.62E+00 & \cellcolor{gray!30}\textbf{7.96E+00} & 2.20E+00 & 3.42E+00 \\
 & $\pm$(3.90E-01) & $\pm$(5.75E+00) & $\pm$(5.00E-02) & $\pm$(3.18E+00) & $\pm$(5.75E+00) & $\pm$(5.44E+00) & $\pm$(3.40E-01) & $\pm$(1.44E+00) & $\pm$(5.75E+00) & $\pm$(5.41E+00) & $\pm$(3.40E-01) & $\pm$(2.70E-01) \\ \hline
\multirow{2}{*}{Rastrigin1} & 1.72E+00 & 1.77E+00 & 1.88E+00 & \cellcolor{gray!30}\textbf{2.36E+00} & 1.72E+00 & 1.77E+00 & \cellcolor{gray!30}\textbf{1.88E+00} & 1.84E+00 & 1.82E+00 & 1.78E+00 & 1.90E+00 & \cellcolor{gray!30}\textbf{3.23E+00} \\
 & $\pm$(2.10E-01) & $\pm$(1.60E-01) & $\pm$(2.80E-01) & $\pm$(9.80E-01) & $\pm$(2.10E-01) & $\pm$(1.60E-01) & $\pm$(2.80E-01) & $\pm$(4.00E-02) & $\pm$(1.10E-01) & $\pm$(1.50E-01) & $\pm$(2.60E-01) & $\pm$(4.30E-01) \\ \hline
\multirow{2}{*}{Rastrigin2} & 2.04E+00 & 2.12E+00 & 2.31E+00 & \cellcolor{gray!30}\textbf{2.55E+00} & 2.46E+00 & 2.52E+00 & 2.48E+00 & \cellcolor{gray!30}\textbf{2.84E+00} & 3.03E+00 & 3.16E+00 & 3.92E+00 & \cellcolor{gray!30}\textbf{4.06E+00} \\
 & $\pm$(3.00E-02) & $\pm$(4.00E-02) & $\pm$(1.30E-01) & $\pm$(2.50E-01) & $\pm$(1.00E-02) & $\pm$(3.00E-02) & $\pm$(7.00E-02) & $\pm$(1.09E+00) & $\pm$(5.80E-01) & $\pm$(4.60E-01) & $\pm$(7.70E-01) & $\pm$(6.80E-01) \\ \hline
\multirow{2}{*}{Rastrigin3} & 1.02E+00 & 1.03E+00 & 1.02E+00 & \cellcolor{gray!30}\textbf{1.10E+00} & 1.04E+00 & 1.06E+00 & 1.06E+00 & \cellcolor{gray!30}\textbf{1.33E+00} & 1.04E+00 & 1.09E+00 & 1.10E+00 & \cellcolor{gray!30}\textbf{1.30E+00} \\
 & $\pm$(1.00E-02) & $\pm$(2.00E-02) & $\pm$(1.00E-02) & $\pm$(1.00E-02) & $\pm$(5.00E-03) & $\pm$(2.60E-02) & $\pm$(2.20E-02) & $\pm$(4.10E-02) & $\pm$(7.00E-03) & $\pm$(6.40E-02) & $\pm$(6.00E-02) & $\pm$(2.27E-01) \\ \hline
\multirow{2}{*}{Rastrigin4} & 1.07E+00 & 1.13E+00 & 1.31E+00 & \cellcolor{gray!30}\textbf{1.59E+00} & 1.07E+00 & 1.14E+00 & 1.32E+00 & \cellcolor{gray!30}\textbf{1.68E+00} & 1.07E+00 & 1.14E+00 & 1.33E+00 & \cellcolor{gray!30}\textbf{1.77E+00} \\
 & $\pm$(4.00E-02) & $\pm$(2.00E-02) & $\pm$(6.00E-02) & $\pm$(2.60E-01) & $\pm$(4.00E-02) & $\pm$(2.00E-02) & $\pm$(1.50E-01) & $\pm$(1.40E-01) & $\pm$(4.00E-02) & $\pm$(2.00E-02) & $\pm$(1.60E-01) & $\pm$(1.50E-01) \\ \hline
\multirow{2}{*}{Rastrigin5} & 1.05E+00 & 1.07E+00 & 1.10E+00 & \cellcolor{gray!30}\textbf{1.43E+00} & 1.05E+00 & 1.13E+00 & 1.16E+00 & \cellcolor{gray!30}\textbf{1.51E+00} & 1.05E+00 & 1.13E+00 & 1.22E+00 & \cellcolor{gray!30}\textbf{1.42E+00} \\
 & $\pm$(6.00E-03) & $\pm$(5.00E-03) & $\pm$(3.00E-02) & $\pm$(6.00E-02) & $\pm$(6.00E-03) & $\pm$(6.00E-02) & $\pm$(3.00E-02) & $\pm$(7.00E-02) & $\pm$(6.00E-03) & $\pm$(6.00E-02) & $\pm$(2.00E-02) & $\pm$(3.00E-02) \\ \hline
\multirow{2}{*}{Rastrigin6} & 1.07E+00 & 1.20E+00 & \cellcolor{gray!30}\textbf{1.61E+00} & 1.50E+00 & 1.11E+00 & 1.25E+00 & \cellcolor{gray!30}\textbf{1.64E+00} & 1.52E+00 & 1.24E+00 & 1.40E+00 & 1.70E+00 & \cellcolor{gray!30}\textbf{1.99E+00} \\
 & $\pm$(2.00E-02) & $\pm$(1.10E-01) & $\pm$(2.90E-01) & $\pm$(3.00E-01) & $\pm$(2.00E-02) & $\pm$(1.10E-01) & $\pm$(2.60E-01) & $\pm$(4.00E-02) & $\pm$(1.10E-01) & $\pm$(3.00E-02) & $\pm$(4.40E-01) & $\pm$(3.80E-01) \\ \hline

\multirow{2}{*}{Avg Rank}
& \multirow{2}{*}{3.54}
& \multirow{2}{*}{2.71}
& \multirow{2}{*}{2}
& \multirow{2}{*}{1.63}
& \multirow{2}{*}{3.5}
& \multirow{2}{*}{2.67}
& \multirow{2}{*}{1.96}
& \multirow{2}{*}{1.79}
& \multirow{2}{*}{3.67}
& \multirow{2}{*}{2.83}
& \multirow{2}{*}{2.13}
& \multirow{2}{*}{1.38}
\\
&    &      &    &     
&      &      &      &     
&      &      &      &     
\\ \hline

\end{tabular}
}
\end{table*}

\begin{figure*}[t]
\centering

\subfigure[\emph{Sphere1} 40D]{
\includegraphics[width=0.23\linewidth]{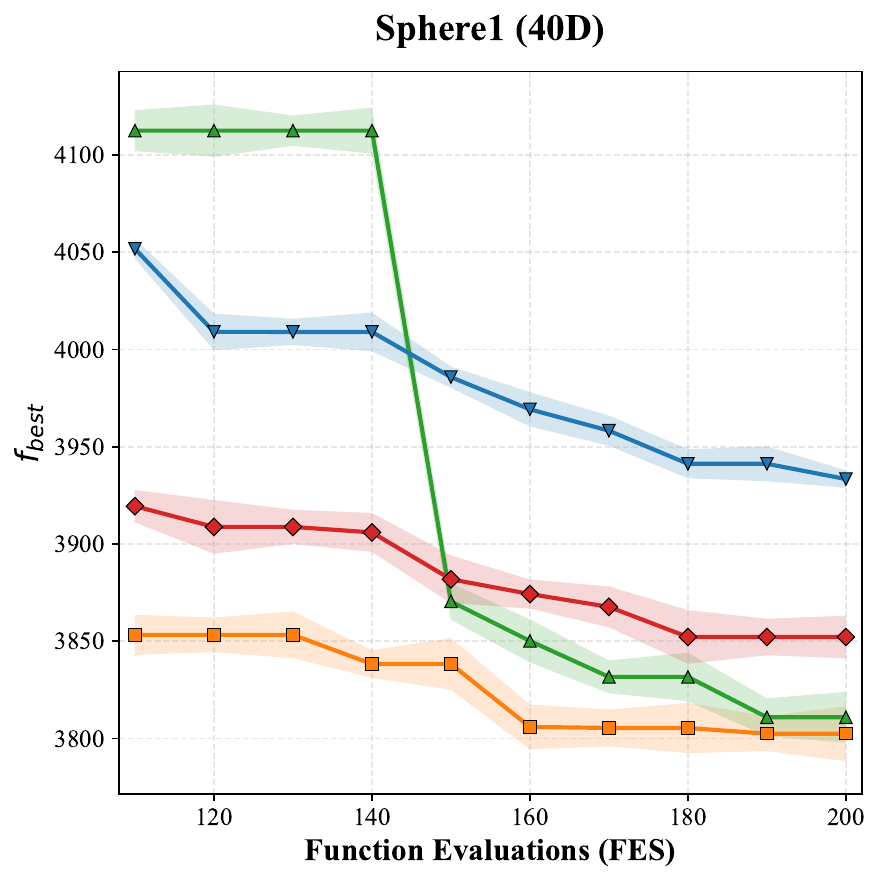}
}
\subfigure[\emph{Ellipsoid1} 40D]{
\includegraphics[width=0.23\linewidth]{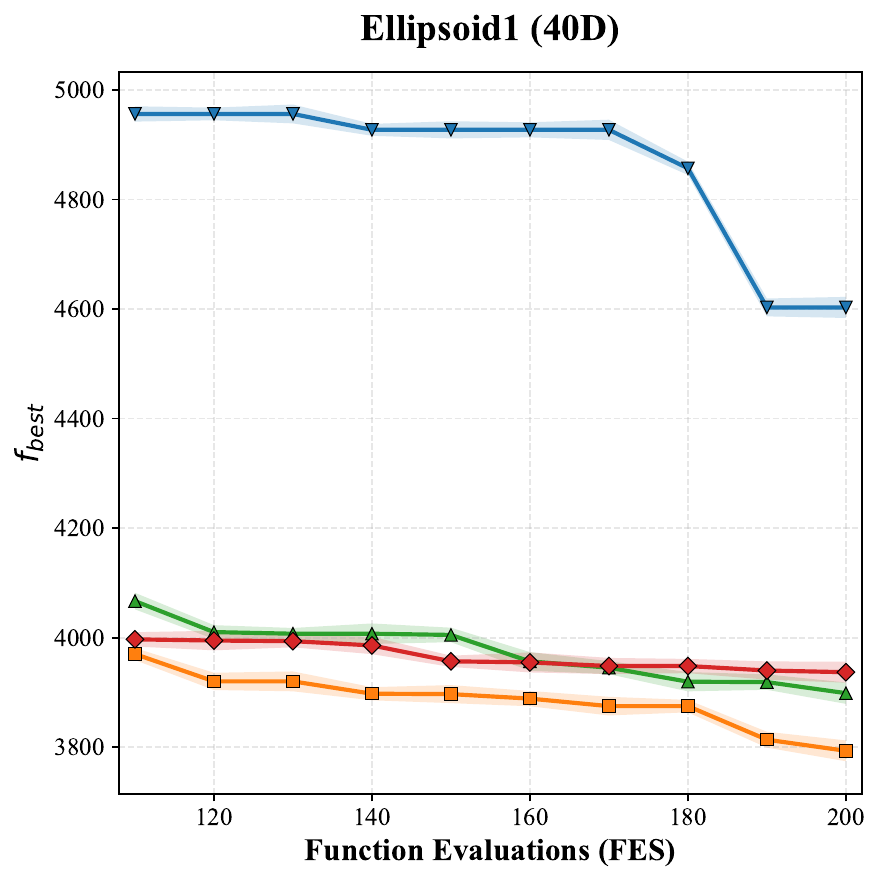}
}
\subfigure[\emph{Bent Cigar1} 40D]{
\includegraphics[width=0.23\linewidth]{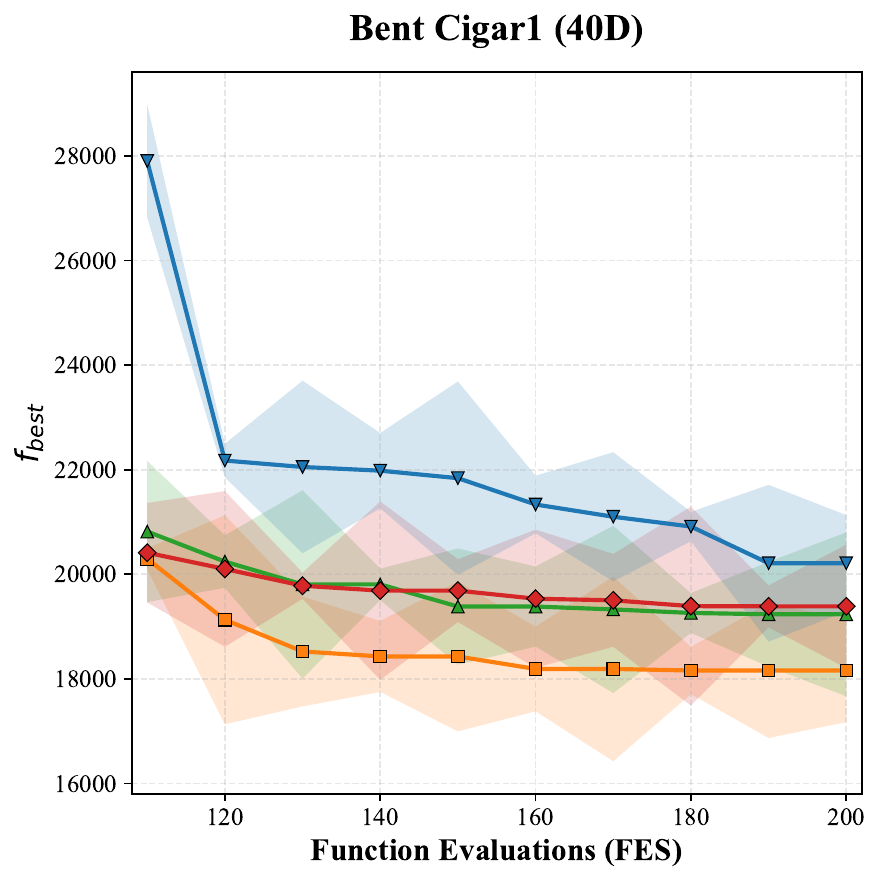}
}
\subfigure[\emph{Rastrigin1} 40D]{
\includegraphics[width=0.23\linewidth]{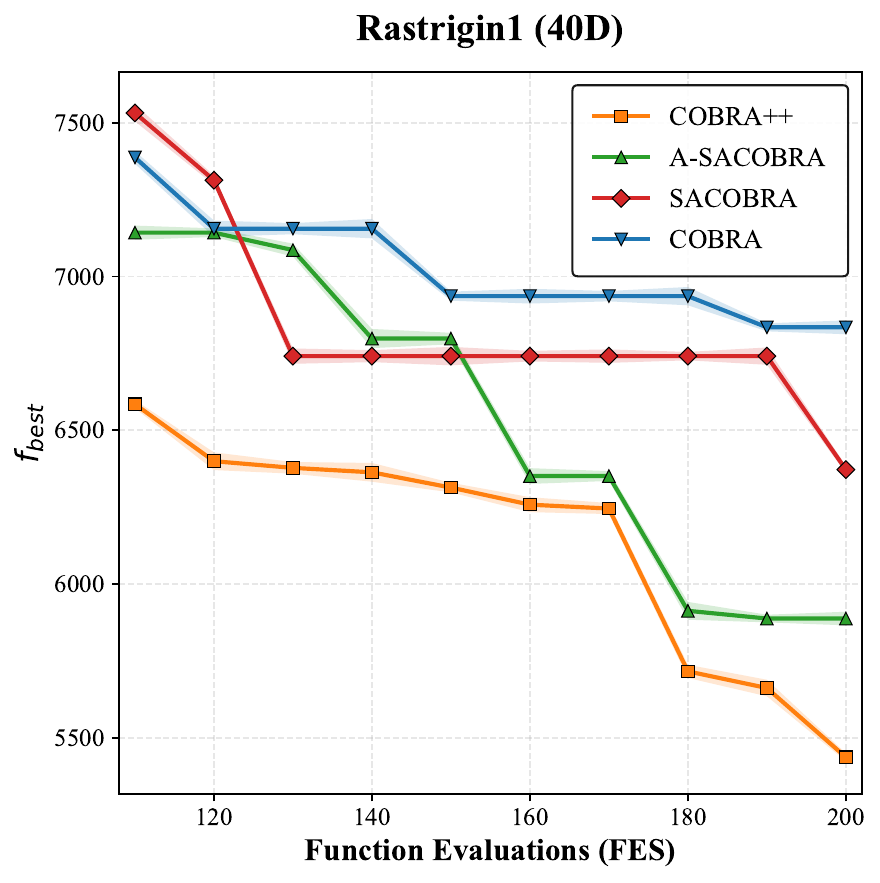}
}
\caption{Performance Comparison on 40D Problems}
\label{fig:40D}

\end{figure*}

\section{Experimental Results}
\label{sec:experiment}
\subsection{Experiment Setup}
\subsubsection{Problem Set}

We evaluate our method on the \textit{bbob-constrained} benchmark from the COCO platform~\cite{hansen2021coco}.
This benchmark comprises 9 basic functions (\textit{Sphere}, \textit{Ellipsoid}, \textit{Bent Cigar}, \textit{Rastrigin}, \textit{Linear Slope}, \textit{Ellipsoid Rotated}, \textit{Discus}, \textit{Different Powers}, \textit{Rotated Rastrigin}), each with 6 instances under different shift, rotation and constraint settings, resulting in a total of $9 \times 6 = 54$ constrained continuous optimization problems.
For the train--test split, we considered all $5 \times 6 = 30$ instances from   \textit{Linear Slope}, \textit{Ellipsoid Rotated}, \textit{Discus}, \textit{Different Powers} as training set, and all $4 \times 6 = 24$ instances from \textit{Sphere}, \textit{Ellipsoid}, \textit{Bent Cigar}, and \textit{Rastrigin} as testing set.
We conduct experiments on problem dimensions 10 and 40.

\subsubsection{Metrics}
To quantify the optimization effectiveness, we adopt a relative improvement~(RI) metric defined below.
Let $f(x^0)$ denotes the objective value of the initial feasible solution, $f(x^\ast)$ denotes the objective value of the true optimum provided by the benchmark, and $f(x^{\text{best}})$ denotes the best objective value obtained by the algorithm.
The metric is defined as
\begin{equation}
\mathrm{RI}=\frac{f(x^0) - f(x^\ast)}{f(x^{\text{best}}) - f(x^\ast)}.
\end{equation}
A larger value of $\mathrm{RI}$ indicates better optimization performance, as it reflects a greater relative improvement achieved from the initial solution toward the optimum.

\subsubsection{Baselines}
We compare COBRA++ with three closely related surrogate-assisted constrained optimization algorithms
to assess the effectiveness of each algorithmic component.
The considered methods are summarized as follows:

\textbf{COBRA}: the original COBRA framework, which serves as the fundamental baseline for surrogate-assisted constrained optimization.
\begin{sloppypar}
\textbf{SACOBRA}: an enhanced variant of COBRA that incorporates self-adaptive mechanisms to strengthen feasibility restoration, thereby improving robustness and overall optimization performance across diverse constrained problem instances.
\end{sloppypar}
\textbf{A-SACOBRA}: a further extension of SACOBRA by maintaining a pool of surrogate models and selecting one surrogate at each iteration based on a heuristic criterion.

All baseline algorithms follow their original implementations and parameter settings, ensuring a fair comparison.

\subsubsection{Settings}
The Q-value module network is implemented as a three-hidden-layers MLP with $\left[32, 64, 128\right]$ hidden neurons and relu activations. All network parameters are initialized using Xavier normal initialization.
In the $\varepsilon$-greedy strategy, $\varepsilon$ is initialized to 1.0, decays by a factor of 0.995 at each step, and is bounded below by 0.01.
We use an experience replay buffer with a capacity of 10{,}000, and the mini-batch size is set to 1024.
The discount factor is set to $\gamma = 0.95$.
The DQN is optimized using the Adam optimizer with a learning rate of $10^{-3}$, and the network parameters are updated every 500 steps.

\textbf{Packages and running platform.}
The implementations of COBRA, SACOBRA and A-SACOBRA are based on the publicly available \texttt{SACOBRA\_Py} library\footnote{\url{https://github.com/WolfgangKonen/SACOBRA_Py}}. The RBF surrogate models are implemented using \texttt{SciPy} library\footnote{\url{https://github.com/scipy/scipy}}.
All experiments were conducted on a computer equipped with an Intel i9-10980XE CPU @ 3.00\,GHz, an NVIDIA RTX 3090 GPU, and 128\,GB of RAM.

\subsection{Generalization Performance Comparison}
Generalization experiment evaluates the optimization performance of algorithms on testing problem set with the same dimension as the training problem set, focusing on whether the learned strategies can effectively adapt to unseen problems.

\subsubsection{10-dimensional Case}
Table~\ref{tab:10d_performance} reports the RI (mean ± standard deviation) of all algorithms on representative 10D problems under 100, 150, and 200 function evaluations, with higher values indicating better optimization performance. The result demonstrates that:

\begin{sloppypar}
(1). \textbf{COBRA vs. SACOBRA}: Overall, SACOBRA surpasses optimization performance of COBRA, achieving better average ranks across all evaluation budgets. This suggests that SACOBRA’s enhancements, including the feasible solution repair strategy and PLOG, significantly improve optimization performance on the testing problems.

(2). \textbf{SACOBRA vs. A-SACOBRA}: 
A-SACOBRA generally achieves higher RI and better average ranks than SACOBRA, indicating that dynamically selecting different surrogate models for different problem instances helps maintain low approximation errors during the optimization process, which in turn leads to improved final optimization performance.

(3). \textbf{A-SACOBRA vs. COBRA++}: 
COBRA++ outperforms A-SACOBRA on most testing problems. This indicates that the RL model selection strategy provides greater flexibility than heuristic-based selection strategy, allowing the algorithm to adapt its surrogate choice according to the optimization and model-specific features rather than relying on predefined strategy. Also, the expanded surrogate model pool offers higher modeling capacity, enabling COBRA++ to better fit diverse landscape structures.
\end{sloppypar}
(4). \textbf{Function-Specific Generalization Challenge}: Although the proposed method achieves near-optimal performance on most testing functions, it fails to find feasible solutions on a few \textit{Bent Cigar} function instances. This may be attributed to the distinctive characteristics of the \textit{Bent Cigar} function, which combines extreme ill-conditioning with strong variable coupling and narrow feasible regions. Such geometric properties are not fully included in the functions in the training set, which may limit the generalization ability of the learned surrogate selection behavior to this specific function family.

\begin{sloppypar}

\subsubsection{40-dimensional Case}
We further evaluate COBRA++ on 40D problems. Due to space limitations, we only present the optimization curves for four representative benchmark functions in Figure~\ref{fig:40D}, where the horizontal axis denotes the number of function evaluations (FEs) and the vertical axis denotes the best objective value found by these algorithms.
Across all four functions, COBRA++ consistently achieves faster convergence and lower final objective values than the baseline methods. These results demonstrate that COBRA++ maintains strong optimization performance and generalization capability in higher-dimensional and more challenging problem settings. The learning-based surrogate model selection strategy and the expanded surrogate model pool enable COBRA++ to effectively adapt to complex high-dimensional landscapes, resulting in robust convergence behavior and superior final optimization performance.
\end{sloppypar}
\begin{table*}[t]
\centering
\caption{Ablation Study Results on 10D Functions}
\label{tab:ablation}
\footnotesize
\resizebox{\textwidth}{!}{%
\begin{tabular}{c|c|cccccccccccc}
\hline
\textbf{Ablation Type} & \textbf{Config.} & \textbf{Sphere1} & \textbf{Sphere2} & \textbf{Sphere3} & \textbf{Sphere4} & \textbf{Sphere5} & \textbf{Sphere6} & \textbf{Ellipsoid1} & \textbf{Ellipsoid2} & \textbf{Ellipsoid3} & \textbf{Ellipsoid4} & \textbf{Ellipsoid5} & \textbf{Ellipsoid6} \\
\hline
\multirow{2}{*}{--} & \multirow{2}{*}{Full}
 & 4.04E+01 & \cellcolor{gray!30}\textbf{8.61E+00} & 7.39E+00 & \cellcolor{gray!30}\textbf{1.03E+02} & \cellcolor{gray!30}\textbf{2.38E+04} & 1.18E+05 & \cellcolor{gray!30}\textbf{5.08E+01} & \cellcolor{gray!30}\textbf{4.09E+01} & 8.11E+00 & 3.98E+00 & \cellcolor{gray!30}\textbf{1.38E+01} & \cellcolor{gray!30}\textbf{1.22E+00} \\
&  & \small($\pm$1.56E+01) & \small($\pm$5.07E+00) & \small($\pm$2.30E+00) & \small($\pm$4.90E+00) & \small($\pm$6.61E+03) & \small($\pm$1.06E+05) & \small($\pm$1.93E+01) & \small($\pm$2.87E+01) & \small($\pm$4.18E+00) & \small($\pm$2.98E+00) & \small($\pm$1.16E+01) & \small($\pm$1.60E-01) \\
\hline
\multirow{2}{*}{\shortstack{Feature\\Extraction}} & \multirow{2}{*}{w/o FE}
 & \cellcolor{gray!30}\textbf{4.21E+01} & 3.12E+00 & \cellcolor{gray!30}\textbf{8.53E+00} & 1.00E+02 & 1.52E+04 & \cellcolor{gray!30}\textbf{4.71E+05} & 4.35E+01 & 1.81E+01 & \cellcolor{gray!30}\textbf{1.23E+01} & \cellcolor{gray!30}\textbf{8.68E+00} & 2.88E+00 & 1.18E+00 \\
&  & \small($\pm$2.45E+00) & \small($\pm$2.00E-02) & \small($\pm$1.32E+00) & \small($\pm$9.21E+00) & \small($\pm$1.51E+04) & \small($\pm$4.48E+05) & \small($\pm$1.91E+01) & \small($\pm$1.60E-01) & \small($\pm$1.40E-01) & \small($\pm$2.88E+00) & \small($\pm$1.42E+00) & \small($\pm$1.80E-01) \\
\hline
\multirow{6}{*}{\shortstack{Surrogate\\Pool}} & \multirow{2}{*}{w/o Cubic}
 & 3.98E+01 & 2.85E+00 & 7.98E+00 & 9.57E+01 & 1.40E+04 & 4.50E+05 & 4.11E+01 & 1.70E+01 & 1.17E+01 & 8.05E+00 & 2.61E+00 & 1.08E+00 \\
&  & \small($\pm$2.21E+00) & \small($\pm$2.00E-02) & \small($\pm$1.18E+00) & \small($\pm$8.55E+00) & \small($\pm$1.39E+04) & \small($\pm$4.32E+05) & \small($\pm$1.82E+01) & \small($\pm$1.40E-01) & \small($\pm$1.20E-01) & \small($\pm$2.65E+00) & \small($\pm$1.32E+00) & \small($\pm$1.60E-01) \\
\cline{2-14}
& \multirow{2}{*}{w/o MQ}
 & 4.05E+01 & 2.90E+00 & 8.10E+00 & 9.72E+01 & 1.45E+04 & 4.58E+05 & 4.18E+01 & 1.73E+01 & 1.19E+01 & 8.20E+00 & 2.70E+00 & 1.10E+00 \\
&  & \small($\pm$2.28E+00) & \small($\pm$2.00E-02) & \small($\pm$1.21E+00) & \small($\pm$8.72E+00) & \small($\pm$1.42E+04) & \small($\pm$4.40E+05) & \small($\pm$1.84E+01) & \small($\pm$1.50E-01) & \small($\pm$1.30E-01) & \small($\pm$2.72E+00) & \small($\pm$1.35E+00) & \small($\pm$1.70E-01) \\
\cline{2-14}
& \multirow{2}{*}{w/o Gauss}
 & 4.09E+01 & 2.93E+00 & 8.20E+00 & 9.80E+01 & 1.47E+04 & 4.61E+05 & 4.21E+01 & 1.74E+01 & 1.20E+01 & 8.30E+00 & 2.71E+00 & 1.11E+00 \\
&  & \small($\pm$2.31E+00) & \small($\pm$2.00E-02) & \small($\pm$1.23E+00) & \small($\pm$8.85E+00) & \small($\pm$1.45E+04) & \small($\pm$4.43E+05) & \small($\pm$1.85E+01) & \small($\pm$1.50E-01) & \small($\pm$1.30E-01) & \small($\pm$2.76E+00) & \small($\pm$1.36E+00) & \small($\pm$1.70E-01) \\
\hline
\hline
\textbf{Ablation Type} & \textbf{Config.} & \textbf{Bent Cigar1} & \textbf{Bent Cigar2} & \textbf{Bent Cigar3} & \textbf{Bent Cigar4} & \textbf{Bent Cigar5} & \textbf{Bent Cigar6} & \textbf{Rastrigin1} & \textbf{Rastrigin2} & \textbf{Rastrigin3} & \textbf{Rastrigin4} & \textbf{Rastrigin5} & \textbf{Rastrigin6}\\
\hline
\multirow{2}{*}{--} & \multirow{2}{*}{Full}
 & \cellcolor{gray!30}\textbf{8.06E+00} & \cellcolor{gray!30}\textbf{1.45E+01} & none & none & \cellcolor{gray!30}\textbf{7.75E+00} & \cellcolor{gray!30}\textbf{6.96E+00} & \cellcolor{gray!30}\textbf{2.36E+00} & \cellcolor{gray!30}\textbf{2.55E+00} & \cellcolor{gray!30}\textbf{1.10E+00} & \cellcolor{gray!30}\textbf{1.59E+00} & \cellcolor{gray!30}\textbf{1.43E+00} & 1.50E+00 \\
&  & \small($\pm$2.83E+00) & \small($\pm$7.09E+00) & -- & --& \small($\pm$3.57E+00) & \small($\pm$3.18E+00) & \small($\pm$9.80E-01) & \small($\pm$2.50E-01) & \small($\pm$1.00E-02) & \small($\pm$2.60E-01) & \small($\pm$6.00E-02) & \small($\pm$3.00E-01) \\
\hline
\multirow{2}{*}{\shortstack{Feature\\Extraction}} & \multirow{2}{*}{w/o FE}
 & 4.05E+00 & 1.02E+01 & \cellcolor{gray!30}\textbf{1.48E+01} & \cellcolor{gray!30}\textbf{3.82E+00} & 2.81E+00 & 1.92E+00 & 2.02E+00 & 2.45E+00 & 1.07E+00 & 1.38E+00 & 1.16E+00 & \cellcolor{gray!30}\textbf{1.72E+00} \\
&  & \small($\pm$1.00E-01) & \small($\pm$1.51E+00) & \small($\pm$2.08E+00) & \small($\pm$2.46E+00) & \small($\pm$1.40E-01) & \small($\pm$5.00E-02) & \small($\pm$2.90E-01) & \small($\pm$1.40E-01) & \small($\pm$1.00E-02) & \small($\pm$6.00E-02) & \small($\pm$3.00E-02) & \small($\pm$3.00E-01) \\
\hline
\multirow{6}{*}{\shortstack{Surrogate\\Pool}} & \multirow{2}{*}{w/o Cubic}
 & 3.78E+00 & 9.58E+00 & 1.39E+01 & 3.52E+00 & 2.58E+00 & 1.76E+00 & 1.88E+00 & 2.25E+00 & 9.80E-01 & 1.25E+00 & 1.05E+00 & 1.56E+00 \\
&  & \small($\pm$9.00E-02) & \small($\pm$1.40E+00) & \small($\pm$1.95E+00) & \small($\pm$2.30E+00) & \small($\pm$1.20E-01) & \small($\pm$5.00E-02) & \small($\pm$2.70E-01) & \small($\pm$1.20E-01) & \small($\pm$1.00E-02) & \small($\pm$5.00E-02) & \small($\pm$3.00E-02) & \small($\pm$2.80E-01) \\
\cline{2-14}
& \multirow{2}{*}{w/o MQ}
 & 3.85E+00 & 9.70E+00 & 1.42E+01 & 3.60E+00 & 2.65E+00 & 1.80E+00 & 1.90E+00 & 2.30E+00 & 1.00E+00 & 1.30E+00 & 1.08E+00 & 1.60E+00 \\
&  & \small($\pm$9.00E-02) & \small($\pm$1.43E+00) & \small($\pm$2.00E+00) & \small($\pm$2.35E+00) & \small($\pm$1.30E-01) & \small($\pm$5.00E-02) & \small($\pm$2.80E-01) & \small($\pm$1.30E-01) & \small($\pm$1.00E-02) & \small($\pm$6.00E-02) & \small($\pm$3.00E-02) & \small($\pm$2.90E-01) \\
\cline{2-14}
& \multirow{2}{*}{w/o Gauss}
 & 3.88E+00 & 9.80E+00 & 1.43E+01 & 3.63E+00 & 2.67E+00 & 1.81E+00 & 1.89E+00 & 2.31E+00 & 1.01E+00 & 1.31E+00 & 1.10E+00 & 1.61E+00 \\
&  & \small($\pm$9.00E-02) & \small($\pm$1.45E+00) & \small($\pm$2.01E+00) & \small($\pm$2.37E+00) & \small($\pm$1.30E-01) & \small($\pm$5.00E-02) & \small($\pm$2.80E-01) & \small($\pm$1.30E-01) & \small($\pm$1.00E-02) & \small($\pm$6.00E-02) & \small($\pm$3.00E-02) & \small($\pm$2.90E-01) \\
\hline
\end{tabular}
}
\begin{tablenotes}
\footnotesize
\item FE = Feature Extraction Module; MQ = Multiquadric; Gauss = Gaussian.
\end{tablenotes}
\end{table*}
\subsection{In-depth Analysis}
In this section, we provide an in-depth analysis to validate the effectiveness of COBRA++ from learning effectiveness and phase-specific model selection behavior.
\subsubsection{Learning Effectiveness}

\begin{figure}[t]
    \centering
    \includegraphics[width=0.5\columnwidth]{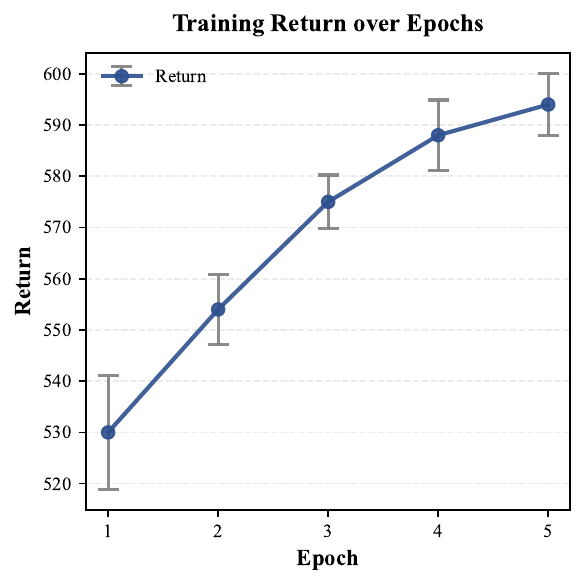}
    \caption{Return curve of COBRA++.}
    \label{fig:return} 
\end{figure}

Figure~\ref{fig:return} shows the training return curve of the model selection policy over epochs. The return steadily increases and converges after five training epochs, indicating that the policy successfully learns to select surrogate models that improve overall optimization performance.

\subsubsection{Phase-Specific Model Selection Behavior Analysis}

\begin{figure}[t]
    \centering
    \includegraphics[width=\columnwidth]{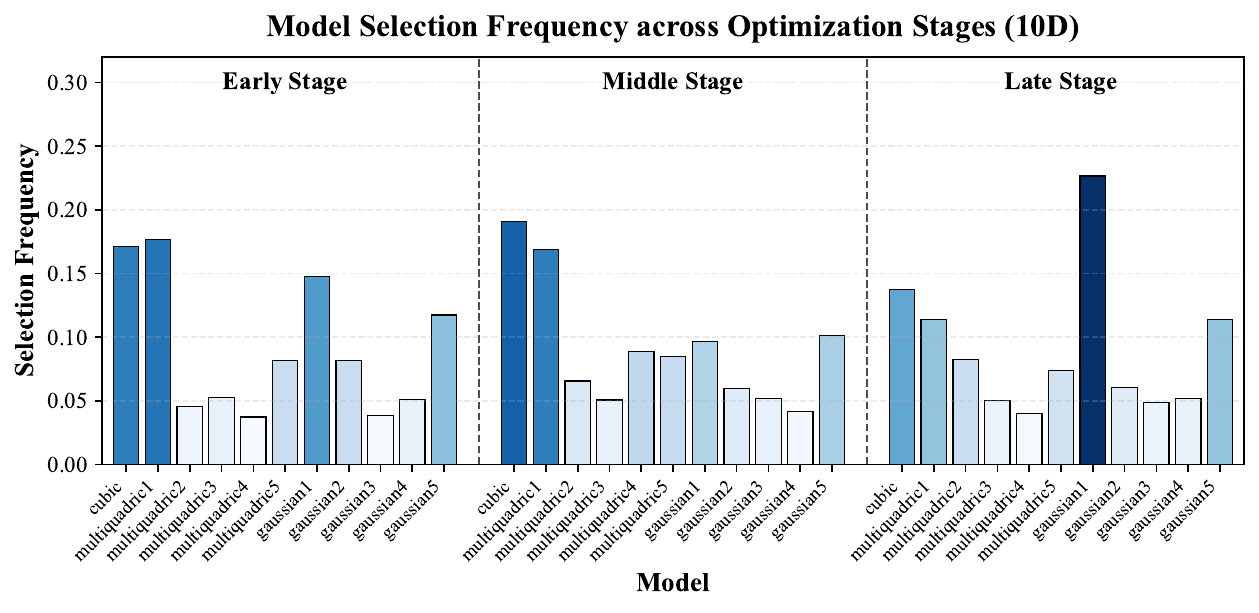}
    \caption{Model selection frequency across three optimization phases}
    \label{fig:frequency} 
\end{figure}

To further investigate how the learned policy selects different models during optimization, we analyze the selection frequency of surrogate models across three optimization phases (early, middle, and late stages) on 10D problems, as shown in Figure~\ref{fig:frequency}. The early, middle, and late phases correspond to the first, middle, and final thirds of the evaluation budget, respectively.

The results reveal distinct phase-dependent model selection behavior. In the early stage, COBRA++ explores different types of surrogate models, while showing a preference for models with narrower kernel widths. This suggests that the policy actively try diverse models types but already favors surrogates with higher local modeling capacity. In the middle stage, cubic and multiquadric RBF models are selected more frequently. These kernels provide higher nonlinearity and flexibility in function approximation, which facilitates global exploration in complex search landscapes. In the late stage, the Gaussian RBF with the smallest kernel width ($w=0.01$) is selected more frequently, reflecting a strong preference for fine-grained local approximation, which is beneficial for refining solutions near the optimum.

\subsection{Ablation Study}

To validate the effectiveness of core design in COBRA++, we conduct ablation experiments in both network architecture design and surrogate model pool design. Specifically, we remove the feature extraction module and selectively remove certain types of surrogate models from the model pool. The results are reported in Table~\ref{tab:ablation}. The ablation experiments is conducted on 10D problems with 100 function evaluations.

\subsubsection{Feature Extraction Module}

The feature extraction module in COBRA++ is designed to enable the Q-value module to capture complex correlations between optimization states and model selections. To evaluate its necessity, we ablate this module by directly concatenating the raw state vectors in an interleaved manner, i.e., $[s_{m,1}^t,\, s_g^t,\, s_{m,2}^t,\, s_g^t, \ldots]$, and feeding them into the Q-value module. As shown in Table~\ref{tab:ablation}, removing the feature extraction module degrades the performance on most problems.
This result highlights the importance of  feature extraction for training the policy in COBRA++. Without feature extraction, the policy is limited to a shallow mapping from raw states to actions, which limits its ability to capture the complex relationships between optimization dynamics and surrogate model performance.

\subsubsection{Surrogate Model Pool}
COBRA++ employs a diverse surrogate model pool consisting of 11 RBF models with different kernel types and width parameters. To assess the importance of model diversity, we conduct ablation experiments by removing each kernel type individually, resulting in reduced model pools.
As shown in Table~\ref{tab:ablation}, removing any kernel type leads to noticeable performance degradation, indicating that each model contributes complementary approximation capabilities. Reducing model pool limits the policy’s flexibility in selecting appropriate surrogates for diverse problem landscapes.

\section{Conclusion}
\label{sec:conclusion}

In this paper, we propose COBRA++, a reinforcement learning-based framework for expensive constrained black-box optimization. To overcome the static surrogate selection policy and limited model diversity in existing COBRA variants, COBRA++ introduces an expanded RBF-based surrogate pool with diverse kernel types and width configurations and an RL-driven model selection policy that dynamically coordinates multiple surrogate models throughout the optimization process. Through comprehensive experiments on the COCO bbob-constrained benchmark, we demonstrate that COBRA++ substantially outperforms vanilla COBRA and its variants across both 10D and 40D problems under strict evaluation budgets. In-depth studies and ablation studies further validate the effectiveness of our design in COBRA++.
As a promising step toward adaptive surrogate-assisted expensive constrained optimization, COBRA++ still has limitations. Future work includes extending to multi-objective scenarios and incorporating diverse surrogate types beyond RBF.



\bibliographystyle{ACM-Reference-Format}
\bibliography{sample-base}

\end{document}